\documentclass[11pt,a4paper]{article}
\usepackage[hyperref]{acl2017}
\usepackage{times}
\usepackage{latexsym}
\usepackage{amssymb}
\usepackage{amsbsy}
\usepackage{graphicx}
\usepackage[utf8]{inputenc}
\usepackage[main=english,bulgarian]{babel}
\usepackage{url}
\usepackage{todonotes}
\usepackage{booktabs}

\aclfinalcopy 

\title{Detecting Toxicity in News Articles: Application to Bulgarian}

\author{Yoan Dinkov, Ivan Koychev\\
  Faculty of Mathematics and Informatics\\
  Sofia University\\
  Sofia, Bulgaria \\
  \texttt{\{jdinkov,koychev\}@uni-sofia.bg} \\\And
  Preslav Nakov \\
  Qatar Computing Research Institute\\
  HBKU\\
  Doha, Qatar\\
  {\tt pnakov@hbku.edu.qa} \\}

\date{}

\begin{document}
\maketitle
\begin{abstract}

Online media aim for reaching ever bigger audience and for attracting ever longer attention span. This competition creates an environment that rewards sensational, fake, and toxic news. To help limit their spread and impact, we propose and develop a news toxicity detector that can recognize various types of toxic content. While previous research primarily focused on English, here we target Bulgarian. We created a new dataset by crawling a website that for five years has been collecting Bulgarian news articles that were manually categorized into eight toxicity groups. Then we trained a multi-class classifier with nine categories: eight toxic and one non-toxic. We experimented with different representations based on ElMo, BERT, and XLM, as well as with a variety of domain-specific features. Due to the small size of our dataset, we created a separate model for each feature type, and we ultimately combined these models into a meta-classifier. The evaluation results show an accuracy of 59.0\% and a macro-F1 score of 39.7\%, which represent sizable improvements over the majority-class baseline (Acc=30.3\%, macro-F1=5.2\%).

\end{abstract}

\section{Introduction}

The number of online news sources has grown dramatically in recent years, and so has the amount of news that has been bombarding users on a daily basis, especially in social media. As people have limited time to spend reading news, capturing people's attention is getting ever harder. Media have to use various techniques to get their readers back such as bigger advertisement and better services. 

\noindent Alternatively, it turns out that an easy way to attract people's attention is to use some toxicity in the articles, as people are intrigued by the unusual.
Thus, our aim here is to try to detect articles that can harm, give false impressions, or deceive the readers. Such articles can use some of the following techniques:
\begin{itemize}
    \item \emph{Sensationalism}: overexposing insignificant or ordinary events by manipulating the main point of an article; 
    \item \emph{Fake news}: news that sound right, but totally misinterpret facts such as statistical data, locations, or dates, with the conscious aim of proving something wrong. 
    \item \emph{Conspiracy theories}: information that usually gives a lot of detail, but does not offer officially confirmed evidence or scientific research to back the claims that are being made. This is typically centered around political, or strange scientific phenomena. 
    \item \emph{Hate speech}: specifically targeting a person or a social group to brutalize them or to bully over the rate of normal conversation, to directly hurt or to manipulate them.
\end{itemize}

Given the proliferation of toxic news online, there have been many efforts to create tools and mechanisms to counteract their effect and spread. Such tools should help preserve and improve the reading integrity. Solving this problem is not a trivial task, and it requires a lot of effort by trained domain experts. Yet, there are limitations in how much it is possible to handle manually in a short period of time (and time is very critical as toxic content spreads fast). Thus, an attractive alternative is to use machine learning and natural language processing to automate the process of toxic news detection.

\noindent While most previous research has focused almost exclusively on English, here we target Bulgarian. In particular, we built a dataset for our experiments based on the knowledge base of \emph{Media Scan}, which has catalogued and characterized many of the Bulgarian online media in the past five years. If a medium published a toxic news, this was recorded and the article, as well as the medium, got labelled accordingly. The analyzed media vary from digital newspapers, to media groups and blogs. For some articles there is detailed explanation with examples about why they were labelled like that. In some cases, the \emph{Media Scan} website describes attempts to contact the authors of an article asking for clarification about some questionable facts that are being reported.

Here we use this information by performing multi-class classification over the \emph{toxicity} labels: \emph{fake news}, \emph{sensations}, \emph{hate speech}, \emph{conspiracies}, \emph{anti-democratic}, \emph{pro-authoritarian}, \emph{defamation}, \emph{delusion}. Note that we allow multiple of these labels simultaneously. We further add a \textit{non-toxic} label for articles that represent good news.


\section{Related Work}\label{sec:related}

The proliferation of false information has attracted a lot of research interest recently. This includes challenging the truthiness of news~\cite{brill2001online,Hardalov2016,Potthast2018}, of news sources~\cite{D18-1389,source:multitask:NAACL:2019,INTERSPEECH2019:youtube}, 
and of social media posts~\cite{Canini:2011,Castillo:2011:ICT:1963405.1963500}, 
as well as studying credibility, influence, bias, and propaganda~\cite{Ba:2016:VERA,Chen:2013:BIW:2492517.2492637,Mihaylov2015FindingOM,Kulkarni:2018:EMNLP,D18-1389,InternetResearchJournal:2018,Barron:19,EMNLP2019:propaganda:finegrained,EMNLP2019:tanbih}. 
Research was facilitated by shared tasks such as the SemEval 2017 and 2019 tasks on Rumor Detection \cite{derczynski-EtAl:2017:SemEval,gorrell-etal-2019-semeval}, the CLEF 2018 and 2019 CheckThat! labs \cite{clef2018checkthat:overall,clef-checkthat:2019,CheckThat:ECIR2019}, which featured tasks on automatic identification \cite{clef2018checkthat:task1,clef-checkthat-T1:2019} and verification \cite{clef2018checkthat:task2,clef-checkthat-T2:2019} of claims in political debates, the FEVER 2018 and 2019 task on Fact Extraction and VERification~\cite{thorne-EtAl:2018:N18-1}, and the SemEval 2019 task on Fact-Checking in Community Question Answering Forums~\cite{mihaylova-etal-2019-semeval}, among others.

\noindent The interested reader can learn more about ``fake news'' from the overview by \newcite{Shu:2017:FND:3137597.3137600}, which adopted a data mining perspective and focused on social media.
Another recent survey  \cite{thorne-vlachos:2018:C18-1} took a fact-checking perspective on ``fake news'' and related problems.
Yet another survey was performed by~\newcite{Li:2016:STD:2897350.2897352}, and it covered truth discovery in general.
Moreover, there were two recent articles in \emph{Science}:
\newcite{Lazer1094} offered a general overview and discussion on the science of ``fake news'', while
\newcite{Vosoughi1146} focused on the  proliferation of true and false news online.

The veracity of information has been studied at different levels:
(\emph{i})~claim (e.g.,~\emph{fact-checking}),
(\emph{ii})~article (e.g.,~\emph{``fake news'' detection}),
(\emph{iii})~user (e.g.,~\emph{hunting for trolls}), and
(\emph{iv})~medium (e.g.,~\emph{source reliability estimation}).
Our primary interest here is at the article-level.

\subsection{Fact-Checking}

At the claim-level, fact-checking and rumor detection have been primarily addressed using information extracted from social media, i.e.,~based on how users comment on the target claim \cite{Canini:2011,Castillo:2011:ICT:1963405.1963500,ma2016detecting,P17-1066,dungs-EtAl:2018:C18-1,kochkina-liakata-zubiaga:2018:C18-1}.
The Web has also been used as a source of information \cite{mukherjee2015leveraging,popat2016credibility,Popat:2017:TLE:3041021.3055133,RANLP2017:factchecking:external,AAAI2018:factchecking,baly-EtAl:2018:N18-2,EMNLP2019:fauxtography}.

In both cases, the most important information sources are
\emph{stance} (does a tweet or a news article agree or disagree with the claim?), and
\emph{source reliability} (do we trust the user who posted the tweet or the medium that published the news article?).
Other important sources are linguistic expression, meta information, and temporal dynamics.

\subsection{Stance Detection}

Stance detection has been addressed as a task in its own right, where models have been developed based on data from the Fake News Challenge \cite{riedel2017simple,thorne-EtAl:2017:NLPmJ,NAACL2018:stance,hanselowski-EtAl:2018:C18-1}, or from SemEval-2017 Task~8 \cite{derczynski-EtAl:2017:SemEval,dungs-EtAl:2018:C18-1}. It has also been studied for other languages such as Arabic \cite{DarwishMZ17,baly-EtAl:2018:N18-2,EMNLP2019:Stance:crosslanguage:contrastive}.

\subsection{Source Reliability Estimation}

Unlike stance detection, the problem of source reliability remains largely under-explored.
In the case of social media, it concerns modeling the user\footnote{User modeling in social media and news community forums has focused on finding malicious users such as opinion manipulation \emph{trolls}, paid \cite{Mihaylov2015ExposingPO} or just perceived \cite{Mihaylov2015FindingOM,mihaylov-nakov:2016:P16-2,InternetResearchJournal:2018,AAAI2018:factchecking}, \emph{sockpuppets} \cite{Maity:2017:DSS:3022198.3026360}, \emph{Internet water army} \cite{Chen:2013:BIW:2492517.2492637}, and \emph{seminar users} \cite{SeminarUsers2017}.} who posted a particular message/tweet, while in the case of the Web, it is about the trustworthiness of the source (the URL domain, the medium).

The source reliability of news media has often been estimated automatically based on the general stance of the target medium with respect to known manually fact-checked claims, without access to gold labels about the overall medium-level factuality of reporting \cite{mukherjee2015leveraging,popat2016credibility,Popat:2017:TLE:3041021.3055133,Popat:2018:CCL:3184558.3186967}.
The assumption is that reliable media agree with true claims and disagree with false ones, while for unreliable media it is mostly the other way around.
The trustworthiness of Web sources has also been studied from a Data Analytics perspective.
For instance, \citet{Dong:2015:KTE:2777598.2777603} proposed that a trustworthy source is one that contains very few false facts.

Note that estimating the reliability of a source is important not only when fact-checking a claim \cite{Popat:2017:TLE:3041021.3055133,DBLP:conf/aaai/NguyenKLW18}, but such reliability scores can be used as an important prior when addressing article-level factuality tasks such as ``fake news'' and click-bait detection \cite{brill2001online,Hardalov2016,RANLP2017:clickbait,desarkar-yang-mukherjee:2018:C18-1,prezrosas-EtAl:2018:C18-1}.

\subsection{``Fake News'' Detection}

Most work on ``fake news'' detection has relied on medium-level labels, which were then assumed to hold for all articles from that source.

\citet{DBLP:journals/corr/HorneA17} analyzed three small datasets ranging from a couple of hundred to a few thousand articles from a  couple of dozen sources, comparing (\emph{i})~real news vs. (\emph{ii})~``fake news'' vs. (\emph{iii})~satire, and found that the latter two have a lot in common across a number of dimensions. They designed a rich set of features that analyze the text of a news article, modeling its complexity, style, and psychological characteristics. 

\noindent They found that ``fake news'' pack a lot of information in the title (as the focus is on users who do not read beyond the title), and use shorter, simpler, and repetitive content in the body (as writing fake information takes a lot of effort). Thus, they argued that the title and the body should be analyzed separately.

In follow-up work, \citet{DBLP:journals/corr/abs-1803-10124} created a large-scale dataset covering 136K articles from 92 sources from \url{opensources.co}, which they characterize based on 130 features from seven categories: structural, sentiment, engagement, topic-dependent, complexity, bias, and morality. We use this set of features when analyzing news articles.

In yet another follow-up work, \citet{Horne:2018:ANL:3184558.3186987} trained a classifier to predict whether a given news article is coming from a reliable or from an unreliable (``\emph{fake news}'' or \emph{conspiracy})\footnote{We show in parentheses, the labels from \url{opensources.co} that are used to define a category.} source. Note that they assumed that all news from a given website would share the same reliability class. Such an assumption is fine for training (distant supervision), but it is problematic for testing, where manual documents-level labels are needed. 

\citet{Potthast2018} used 1,627 articles from nine sources, whose factuality has been manually verified by professional journalists from BuzzFeed. They applied stylometric analysis, which was originally designed for authorship verification, to predict factuality (fake vs. real).

\citet{rashkin-EtAl:2017:EMNLP2017} focused on the language used by ``fake news'' and compared the prevalence of several features in articles coming from trusted sources vs. hoaxes vs. satire vs. propaganda. However, their linguistic analysis and their automatic classification were at the article level and they only covered eight news media sources.

\subsection{Work for Bulgarian}

We are aware of only one piece of previous work for Bulgarian that targets toxicity. In particular, \cite{RANLP2017:clickbait} built a fake news and click-bait detector for Bulgarian based on data from a hackaton.

While most of the above research has focused on isolated and specific task (such as trustworthiness, fake news, fact-checking), here we try to create a holistic approach by exploring several toxic and non-toxic labels simultaneously.

\begin{table}[tbh]
    \begin{center}
    \begin{tabular}{lr}
    \toprule
    \bf Characteristic & \bf Value \\ 
    \midrule
    Toxic articles                    &  221 \\
    Non-toxic articles                  &  96 \\
    Media                              &  164 \\
    Average title length (chars)    &  70.47 \\
    Average title length (words)       &  11.08 \\
    Average text length (chars)       &  3,613.70 \\
    Average text length (words)          &  556.83 \\
    Average text length (sentences)     &  31.64 \\
    \bottomrule
    \end{tabular}
    \end{center}
    \caption {Statistics about the dataset.\label{tab:dataset_information}}
\end{table}

\begin{figure}[tbh]
  \centering
  \includegraphics[width=8cm]{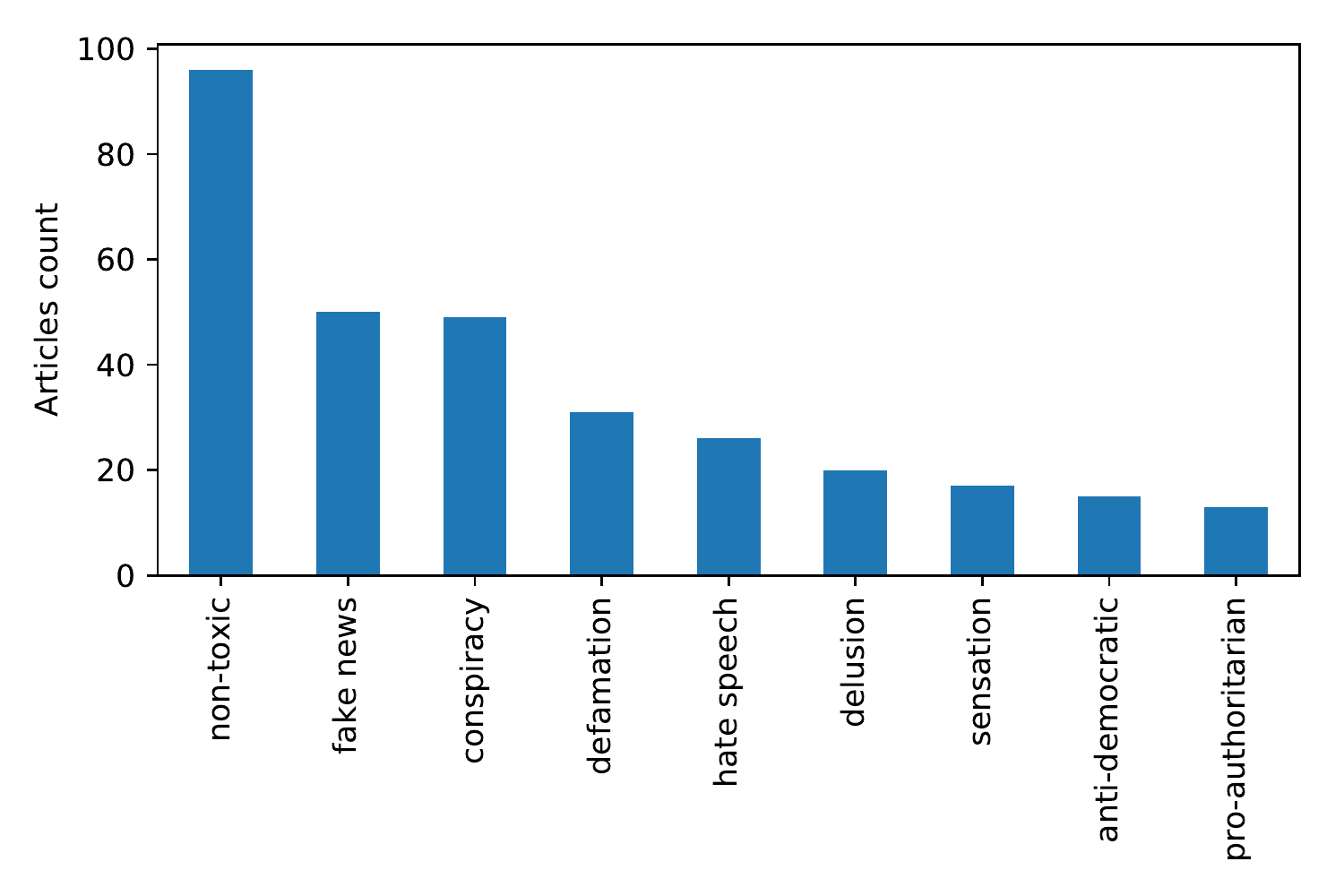}
  \caption{Label distribution in the dataset.\label{fig:labels_distribution}}
\end{figure}

\section{Data}
\label{sec:data}

We used \emph{Media Scan}\footnote{\url{http://mediascan.gadjokov.com}} as a source of toxicity labels for Bulgarian media Web sites. The site contains information about 700 media, and 150 of them are associated with at least one toxic article. Many of these toxic articles are removed after the respective media have been contacted and informed about the problems with these articles. Naturally, the author of \emph{Media Scan} wanted to preserve the original form of the evaluated article, and thus had a link to a PDF copy in case the original HTML page was not accessible. We only crawled the HTML for articles that have not been changed or removed at the time we created the dataset.

For each Web page with a toxic label, we ran a mechanical crawler to obtain its contents. This was not very reliable as each individual medium site has its own structure, while the crawler expected more or less semantic and valid HTML to be able to process it. Thus, we manually verified the data, fixed any issues we could find and added any missing information. We ended up with a little over 200 articles with some kind of toxicity. 

\noindent In addition to this dataset of only toxic articles, we added some ``non-toxic'' articles, fetched from media without toxicity examples in Media Scan: we added a total of 96 articles from 25 media.

Table \ref{tab:dataset_information} shows statistics about the dataset, and Figure \ref{fig:labels_distribution} shows the distribution of the labels.

\section{Method}
\label{sec:method}

We used a feature-rich classifier based on logistic regression and a neural network.

For each article, we extracted its title and its body. We further extracted some meta information about the corresponding news medium. As some NLP resources are only available or are better for English, we translated the articles to English, by using Google Translate API, so that we can extract features from them as explained in subsections \ref{sub:sec:bert}, \ref{sub:sec:use},
\ref{sub:sec:elmo},
and \ref{sub:sec:nela}.

\subsection{LSA}
\label{sub:sec:lsa}

We trained a Latent Semantic Analysis (LSA) model on our data. We first built TF.IDF vectors for the title and the body. Then, we applied singular value decomposition (SVD) to generate vectors of 15 dimensions for the titles and of 200 dimensions for the article bodies.

\subsection{BERT}
\label{sub:sec:bert}

We used BERT \cite{devlin2018bert} for sentence representation, which has achieved very strong results on eleven natural language processing tasks including GLUE, MultiNLI, and SQuAD. Since then, it was used to improve over the state of the art for a number of NLP tasks. The original model was trained on English Wikipedia articles (2500M words). Due to the model complexity and to its size,  it is hard to find enough data that represents a specific domain for a specific language.

We used BERT-as-a-service, which generates a vector of 768 numerical values for a given text. In its original form, this is a sentence representation tool, but we used it to generate text over the first 512 tokens of our article's title or text. We used the multilingual cased pretrained model\footnote{\url{http://github.com/google-research/bert\#pre-trained-models}}. We experimented with all possible pooling strategies for representing the article title and its body, and we eventually chose the following pooling strategies: \textit{REDUCE\_MAX} for the title and \textit{CLS\_TOKEN} for the text of the article.

\subsection{Stylometric Features}
For the title and the body of each article, we calculate the following features:

\begin{itemize}
    \item \textit{avg\_word\_length\_title}:  average length of the word in the article title;
    \item \textit{avg\_word\_length\_text}: average length of the words in the article body
    \item \textit{word\_count\_title}: number of words in the article title;
    \item \textit{word\_count\_text}: number of words in the article body;
    \item \textit{char\_count\_title}: number of characters in the article title;
    \item \textit{char\_count\_text}: number of characters in the article body;
    \item \textit{spec\_char\_count\_title}: number of specific (non-alpha-numeric) characters in the article title;
    \item \textit{spec\_char\_count\_text}: number of specific (non-alpha-numeric) characters in the article body;
    \item \textit{upper\_char\_count\_title}: number of uppercase characters in the article title;
    \item \textit{upper\_char\_count\_text}: number of uppercase characters in the article body;
    \item \textit{upper\_word\_count\_title}: number of words starting with an uppercase character in the article title;
    \item \textit{upper\_word\_count\_text}: number of words starting with an uppercase character in the article body;
    \item \textit{sentence\_count\_text}: number of sentences in the article;
    \item \textit{avg\_sentence\_length\_char\_text}: average length of the sentences in the article body, in terms of characters;
    \item \textit{avg\_sentence\_length\_word\_text}: average length of the sentences in the article body, in terms of words;
\end{itemize}

\subsection{Media Features}
\label{sub:sec:media_meta}

We further extracted binary and numerical features characterizing the medium the article came from:

\begin{itemize}
    \item \textit{editor}: its value is 1 if the target medium has a designated chief editor, and it is 0 otherwise;
    \item \textit{responsible\_person}: its value is 1 if the target medium has a responsible person, and it is 0 otherwise;
    \item \textit{bg\_server}: its value is 1 if the target medium's location is in Bulgaria, and it is 0 otherwise;
    \item \textit{popularity}: reciprocal value of the target medium's rank the Web traffic analysis platform Alexa\footnote{\url{http://www.alexa.com/}};
    \item \textit{domain\_person}: its value is 1 if the target medium has a designated owner, and it is 0 otherwise;
    \item \textit{days\_existing}: number of days between when the medium was created and 01.01.2019. As this value is quite large, we take the logarithm thereof. For example, a medium created on Januarty 1, 2005 would have 5,113 days of existence, which would correspond to 3.70 as this feature's value.
\end{itemize}

\subsection{XLM}
\label{sub:sec:xlm}

We further used cross-lingual representations from the Facebook's XLM model \cite{conneau2019xlm}, which creates cross-lingual representations based on the Transformer, similarly to BERT. This model is pretrained for 15 languages including Bulgarian and English. We use their pretrained models, which were fine-tuned for Cross-lingual Natural Language Interence (XNLI) tasks. This yielded a 1024-demnsional representation for the title, and another one for the article body.

\subsection{Universal Sentence Encoder}
\label{sub:sec:use}

We also extracted representation using Google's Universal Sentence Encoder, or USE, \cite{cer2018universal}. We used the pretrained model from TF Hub\footnote{\url{http://tfhub.dev/google/universal-sentence-encoder/2}}. As the model is only available for English, we used the translations of the news articles. We passed the model the first 300 tokens for each title or body to generate 512-dimensional vectors.

\subsection{ElMo}
\label{sub:sec:elmo}
Next, we use deep contextualized word representations from ElMo, which uses generative bidirectional language model pre-training \cite{peters2018deep}. The model yields 1024-dimensional representation, which we generate separately for the article title and for its body.

\subsection{NELA Features}
\label{sub:sec:nela}

Finally, we use features from the NELA toolkit \cite{Horne:2018:ANL:3184558.3186987}, which were previously shown useful for detecting fake news, political bias, etc. The toolkit implements 129 features, which we extract separately for the article title and for its body:

\begin{itemize}
\item \textbf{Structure}: POS tags, linguistic features (function words, pronouns, etc.), and features for clickbait title classification from \cite{clickbait:2016};
\item \textbf{Sentiment}: sentiment scores using lexicons \citep{recasens2013linguistic,mitchell2013geography} and full systems \cite{gilbert2014vader};
\item \textbf{Topic}: lexicon features to differentiate between science topics and personal concerns;
\item \textbf{Complexity}: type-token ratio, readability, number of cognitive process words (identifying discrepancy, insight, certainty, etc.);
\item \textbf{Bias}: features modeling bias~\citep{recasens2013linguistic,mukherjee2015leveraging} and subjectivity~\cite{horne2017identifying};
\item \textbf{Morality}: features based on the Moral Foundation Theory \cite{graham2009liberals} and lexicons \cite{lin2017acquiring}
\end{itemize}

A summary of all features is shown in Table \ref{tab:features}.

\begin{table}[h]    
    \centering
    \begin{tabular}{lrr}
    \toprule
    \textbf{Feature Group}  &\textbf{Title} &\textbf{Body}\\ 
    \midrule
    BERT & 768 & 768\\
    ElMO & 1,024 & 1,024 \\ 
    LSA  & 15 & 200\\
    NELA & 129 & 129 \\
    Stylometry & 19 & 6\\
    USE  & 512 & 512 \\
    XLM & 1,024 & 1,024 \\ 
    \hline
    Media & \multicolumn{2}{c}{6} \\
    \bottomrule
    \end{tabular}
    \caption {\label{tab:features}Summary of our features.}
\end{table}

\section{Experiments and Evaluation}

\subsection{Experimental Setup}

We used logistic regression as our main classification method.
As we have a small dataset, we performed 5-fold cross-validation.
For evaluation, we used accuracy and macro-average F$_1$ score. The results are presented in Table~\ref{tab:feature_comparison}.

Our baseline approach is based on selecting the most frequent class, i.e.,~\textit{non-toxic}, which covers 30.30\% of the dataset (see Table \ref{tab:dataset_information}).

\subsection{Individual Models}

We evaluated a total of 14 setups for feature combination. Four of them represent features generated from the original article's title and body as well as a combination thereof. The next four setups present feature sets generated from the English translation as well as a combination thereof. The final section of Table~\ref{tab:feature_comparison} shows three setups that are somewhat language-independent: meta media, all features combined together as well as a meta classifier. We tuned the logistic regression for each individual experimental setup, using an additional internal cross-validation for the training part of each experiment in the 5-fold cross-validation. In total, 15,000 additional experiments have been conducted to complete the fine-tuning.

We can see in Table \ref{tab:feature_comparison} that the BERT features (setups 2, 9) perform well both for English and for Bulgarian. The feature combinations (setups 6, 11, 13) do not yield good results as this increases the number of features, while the number of training examples remains limited. Using only the right 6 meta features about the target medium yields 12\% improvement over the baseline. Interestingly, LSA turns out to be the best text representation model. 

\begin{table*}[tbh]
\centering
\begin{tabular}{lclrrr}
\toprule
\textbf{Language} &\textbf {$N$} &\textbf{Feature Set}  &\textbf{Dimension}  &\textbf{Accuracy}  &\textbf{F1-macro}\\
\midrule
-  & 1 & Baseline & -                       & 30.30  &  05.17\\ 
\midrule
BG & 2 & BERT(title), BERT(text)            & 1,536  & 47.69  &  32.58\\ 
   & 3 & XLM(title), XLM(text)              & 2,048  & 38.50  &  24.58\\ 
   & 4 & Styl(title), Styl(text)            & 15    & 31.89  &  08.51\\ 
   & 5 & LSA(title), LSA(text)              & 215   & 55.59  &  42.11\\ 
   & 6 & Bulgarian combined                 & 3,824  & 39.43  &  24.38\\ 
   \midrule 
EN & 7 & USE(title), USE(text)              & 1,024  & 53.70  &  40.68\\
   & 8 & NELA(title), NELA(text)            & 258   & 36.36  &  23.04\\
   & 9 & BERT(title), BERT(text)            & 1,536  & 52.05  &  39.78\\
   & 10 & ElMO(title), ElMO(text)           & 2,048  & 54.60  &  40.95\\
   & 11  & English  combined                & 4,878  & 45.45  &  31.42\\ 
   \midrule
-  & 12 & Media meta                        & 6     & 42.04  &  15.64\\
   & 13 & All combined                      & 8,694  & 38.16  &  26.04\\
   & 14 & \textbf{Meta classifier}          & \textbf{153}   & \textbf{59.06}  &  \textbf{39.70}  \\
   \bottomrule
\end{tabular}
\caption{Evaluation results. \label{tab:feature_comparison}}
\end{table*}

\subsection {Meta Classifier}
\label{sub:sec:meta_classifier}

Next we tried a meta classifier. For this purpose, we extracted the posterior probabilities of the individual classifiers in Table~\ref{tab:feature_comparison} (2-5, 7-10, 12). Then, we trained a logistic classifier on these posteriors (we made sure that we do not leak information about the labels when training the meta classifier).

We can see in Table~\ref{tab:feature_comparison} that the meta classifier yielded the best results, outperforming all of the individual models, and achieving 3.5\% absolute gain in terms of accuracy.

\subsection{Analysis}

\label{sub:sec:additional_evaluation}

\begin{figure}[h]
  \includegraphics[width=9.3cm]{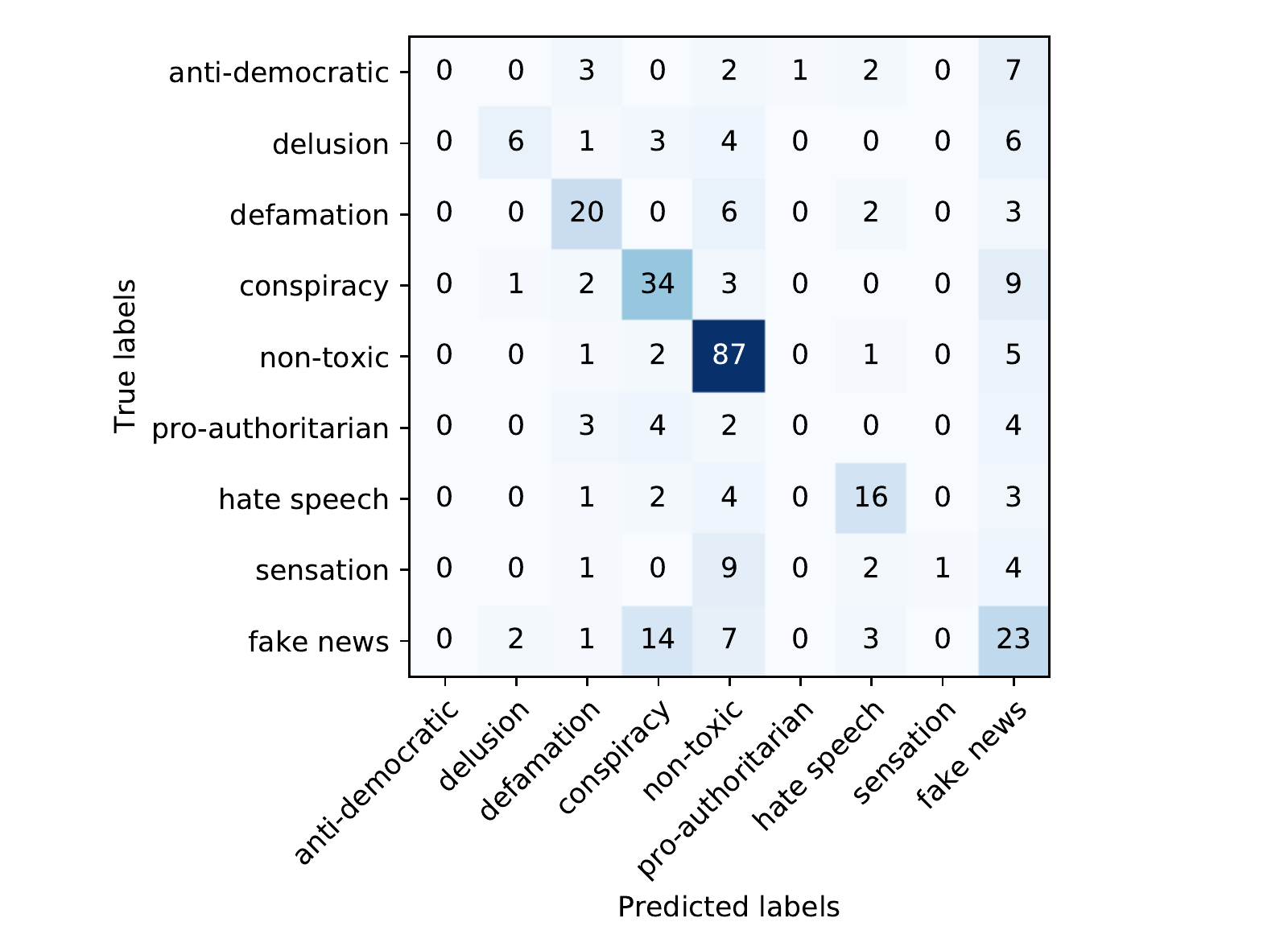}
  \caption{\label{fig:confusion_matrix}Confusion matrix.}
\end{figure}

Figure \ref{fig:confusion_matrix} shows a confusion matrix over the entire dataset for the best model in Table~\ref{tab:feature_comparison}, namely experiment 14. We can see that the model works best for the biggest \textit{non-toxic} class. A decent chunk of \textit{fake news} samples are misclassified as \textit{conspiracy} as those two classes are the second and the third largest ones. For three of the labels, there are hardly any predictions; these are the smallest classes, and three of them combined cover less than 18\% of the dataset.

\begin{figure}[h]
  \centering
  \includegraphics[width=7cm]{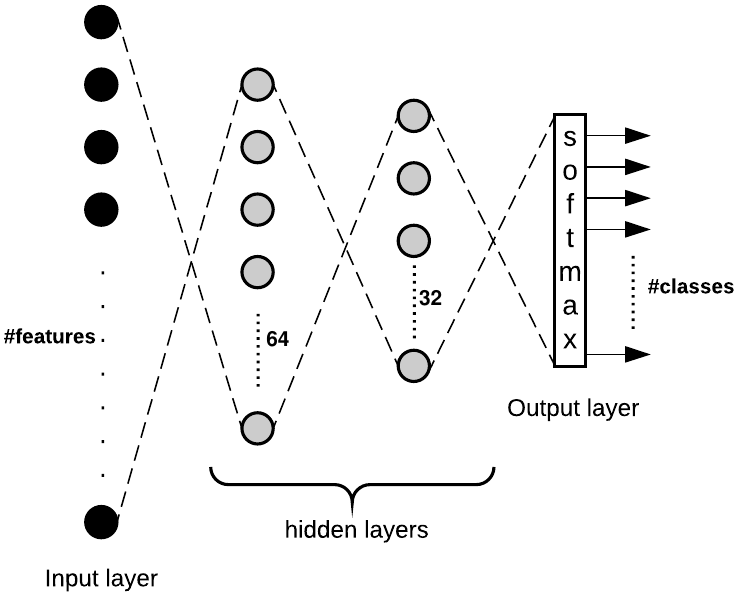}
  \caption{\label{fig:nn}Our neural network.}
\end{figure}

\subsection {What Did Not Work}

\paragraph{Oversampling} We evaluated all models from Table~\ref{tab:feature_comparison} but with oversampling the small classes. We tried simple random oversampling as well as the more complex SMOTE \cite{chawla2002smote} model. None of these techniques yielded improvements.

\paragraph{Neural Network} We also tried a feed forward neural network with two hidden densely connected layers, with 64 nodes (ReLU activation) and 32 nodes (Tanh activation), and a dropout rate of 0.35 for each of them; see Figure~\ref{fig:nn}.
We used Adam for optimization, and we tried various parameter values, but we could not get improvements, possibly due to the small size of our dataset.

\section{Discussion}

Below we compare the performance of the models when working with Bulgarian vs. English resources, and we further discuss issues related to the small size of our dataset.

\subsection{Language-Related Issues}

The first part of the feature comparison in Table \ref{tab:feature_comparison} is between English and Bulgarian, and it is interesting to compare them. The first comparison is between the BERT features. Even though we used a pre-trained BERT model, with same pooling techniques there is close to 4.5\% absolute improvement when using the English translation. This is probably due to the English BERT being trained on more data as the English Wikipedia is much bigger for English: 5.7M English articles vs. just 250K Bulgarian articles.

Another notable comparison is between the types of models. Two of the Bulgarian feature sets are created via local models (experiments 4 and 5), while all of the English experiments are from transfer-learning. We can see that LSA (experiment 5) is the best feature set, and one can argue that this is to be expected. On such a small dataset in a non-English language, it is hard to represent the text with pre-trained models. Nevertheless, we can see that a combination between only pre-trained models (experiment 11) performs better compared to fusion between local and transfer-learning models (experiment 6).

\subsection{Data Size Issues}

There are several aspects of the above experiments where we can observe the negative effect of having insufficient data. First, in experiments 6, 11, 12 in Table~\ref{tab:feature_comparison}, we can see that the combination of features performs worse in each language group compared to single feature types. 

Another place where we felt we had insufficient data was in the neural network experiments, where we had many more parameters to train than in the simple logistic regression. 

A related problem is that of class imbalance: we have seen in Section \ref{sub:sec:additional_evaluation} above that the smallest classes were effectively ignored even by our best classifier. We can see in Figure~\ref{fig:labels_distribution} that those three classes have less than 60 articles combined, while the ``non-toxic'' only had 96 articles.

\section{Conclusion and Future Work}

We have presented experiments in detecting the toxicity of news articles. 
While previous research was mostly limited to English, here we focused on Bulgarian. We created a new dataset by crawling a website that has been collecting Bulgarian news articles and manually categorized them in eight toxicity groups. 
We then trained a multi-class classifier with nine categories: eight toxic and one non-toxic. We experimented with a variety of representations based on ElMo, BERT, xand XLM. We further used a variety of domain-specific features, which we eventually combined in a meta classifier. The evaluation results show an accuracy of 59.0\% and a macro-F1 score of 39.7\%, which represent sizable improvements over the majority-class baseline (Acc=30.3\%, macro-F1=5.2\%).

In future work, we plan to extend and also to balance the dataset. This can be achieved by either exploring another source for articles using the methodology of \emph{Media Scan}, or by processing the unstructured PDF article, which we ignored in the present study.
We also plan to explore new information sources. 
From a technical point of view, we would like to improve the neural network architecture as well as the oversampling techniques (with possible combination with undersampling).

\paragraph{Data and Code} We are releasing all of the code for our experiments in a public repository that can be found in GitHub\footnote{\url{github.com/yoandinkov/ranlp-2019}} with explanations about how to reproduce our environment. In that repository, we further release the full dataset together with the generated features, all the textual data, all the translations and all the meta data about the articles. 

\section*{Acknowledgements}

This research is part of the Tanbih project,\footnote{\url{http://tanbih.qcri.org/}} which aims to limit the effect of ``fake news'', propaganda and media bias by making users aware of what they are reading. The project is developed in collaboration between the Qatar Computing Research Institute (QCRI), HBKU and the MIT Computer Science and Artificial Intelligence Laboratory (CSAIL).

This research is also partially supported by Project UNITe BG05M2OP001-1.001-0004 funded by the OP ``Science and Education for Smart Growth'' and the EU via the ESI Funds.

\bibliography{acl2017}
\bibliographystyle{acl_natbib}

\end{document}